\documentclass[letterpaper, 10 pt, conference]{ieeeconf} 

\IEEEoverridecommandlockouts 

\usepackage{comment}
\usepackage[dvipsnames]{xcolor}
\usepackage{hyperref}
\usepackage{graphicx}

\newcommand{\dbname}{Med-MPD}

\title{\LARGE \bf
A Clinical Dataset for the Evaluation of Motion Planners in Medical Applications
}

\author{Inbar Fried$^{1,2}$, Jason A. Akulian$^{3}$, and Ron Alterovitz$^{1}$
\thanks{This research was supported by the U.S. National Institutes of Health (NIH) under awards R01EB024864 and F30CA265234, and by the National Science Foundation (NSF) under awards 2008475 and 2038855.}
\thanks{The authors acknowledge the National Cancer Institute and the Foundation for the National Institutes of Health, and their critical role in the creation of the free publicly available LIDC/IDRI Database used in this study.}
\thanks{The MR brain images from healthy volunteers used in this paper were collected and made available by the CASILab at The University of North Carolina at Chapel Hill and were distributed by the MIDAS Data Server at Kitware, Inc.}
\thanks{$^{1}$ I. Fried and R. Alterovitz are with the Department of Computer Science, University of North Carolina at Chapel Hill, Chapel Hill, NC 27599, USA. {\tt\footnotesize \{ifried01, ron\}@cs.unc.edu}}
\thanks{$^{2}$ I. Fried is also with the Medical Scientist Training Program, University of North Carolina School of Medicine, Chapel Hill, NC, 27599, USA.}
\thanks{$^{3}$J. A. Akulian is with the Division of Pulmonary Diseases and Critical Care Medicine at the University of North Carolina at Chapel Hill, NC 27599, USA. {\tt\footnotesize jason\_akulian@med.unc.edu}}%
}

\begin{document}

\maketitle
\thispagestyle{empty}
\pagestyle{empty}

\begin{abstract}

The prospect of using autonomous robots to enhance the capabilities of physicians and enable novel procedures has led to considerable efforts in developing medical robots and incorporating autonomous capabilities.
Motion planning is a core component for any such system working in an environment that demands near perfect levels of safety, reliability, and precision.
Despite the extensive and promising work that has gone into developing motion planners for medical robots, a standardized and clinically-meaningful way to compare existing algorithms and evaluate novel planners and robots is not well established.
We present the Medical Motion Planning Dataset (\dbname), a publicly-available dataset of real clinical scenarios in various organs for the purpose of evaluating motion planners for minimally-invasive medical robots.
Our goal is that this dataset serve as a first step towards creating a larger robust medical motion planning benchmark framework, advance research into medical motion planners, and lift some of the burden of generating medical evaluation data.

\end{abstract}
\section{Introduction}

Automation of medical robots for clinical procedures or subtasks is increasingly being shown to be feasible.
Achieving autonomy in interventional medical procedures has a lot of potential benefits for patient care and hospital efficiency.
Much like teleoperated medical robots, such as the da Vinci (Intuitive Surgical Inc., Sunnyvale, CA), can compensate for physician fatigue and hand instability, autonomous medical robotics can further improve and standardize patient care by accounting for inter- and intra-physician variability while also focusing the physician’s time on sub-tasks that require their expertise. 
However, beyond the technical challenges that exist in hardware and software, a critical, if not the most important, challenge in these systems is making them safe and reliable.
To address these challenges and still benefit from the advantages of automation, integrating motion planning into medical robots to ensure safe motions is essential.

One class of medical robots that has been studied extensively over the past couple decades has been medical continuum robots, which include, for example, concentric tube robots and steerable needles~\cite{webster2010design}.
Many mechanical designs have been proposed for these devices, but at their core, medical continuum robots can follow curvilinear trajectories in 3D, allowing them to curve around obstacles and access regions of the anatomy that are otherwise inaccessible when using straight rigid tools.
The potential benefit of these devices has been proposed in numerous organs and for various medical procedures.
The complex kinematics of these devices in conjunction with the precision required for safe medical procedures make manual operation of these devices unintuitive and impractical.
To overcome this challenge, autonomous robots have been proposed that actuate the medical continuum robot following a planned trajectory.

Despite the numerous motion planners that have been proposed for medical continuum robots, to the best of our knowledge, a benchmarking dataset to evaluate the performance of these algorithms does not exist.
The lack of a shared benchmarking resource has lead each research group to generate their own testing data, which is often a time intensive effort.
Since the motion planners have been tested in various organs,
in different anatomical models of those organs, and likely with different obstacle resolutions, it is difficult to properly assess the benefits and drawbacks of each proposed motion planning approach and to compare motion planners.
To help evaluate the benefits of robot automation in medicine, it is important to have benchmarks that can robustly and equitably evaluate the performance of algorithms in clinically relevant scenarios.

In this work, we propose \dbname, a medical benchmarking dataset consisting of real clinical motion planning environments
for assessing motion planners for medical continuum robots and related minimally-invasive medical robots. 
The data includes benchmark scenarios defined by the relevant anatomy and the clinical problem in the lungs, liver, and brain.
We make \dbname~publicly available at~\url{https://github.com/UNC-Robotics/Med-MPD}.
\section{Related Work}

There are several robotics datasets and benchmarking suites that have been developed specifically to allow robust evaluation of motion planners~\cite{chamzas2021motionbenchmaker, moll2015benchmarking, heiden2021bench, kastner2022arena} (and citations within).
These works focus on non-medical robots.
There have also been several medical robotics datasets that have been published, but these are mostly focused on computer vision problems like tool or anatomy segmentation, physician training or assessment, and object manipulation, but not on motion planning~\cite{gao2014jhu, mountney2010three, ahmidi2017dataset, madapana2019desk}.
Several works have proposed simulators for medical robots~\cite{chentanez2009interactive, jianu2022cathsim, dreyfus2022simulation}, but there is no set benchmarking dataset with which to compare different motion planning algorithms.

A variety of algorithms for medical continuum robots have been proposed that encompass various organs and clinical applications.
Within the broad class of medical continuum robots, there has been substantial work in developing motion planners for steerable needles in the lungs~\cite{Kuntz2015_IROS, Hoelscher2021_RAL, fu2022resolution}, liver~\cite{adebar2015methods, Liu2016_RAL}, prostate~\cite{xu2008motion, berg2010lqg, Patil2010_BioRob, bernardes20143d}, and brain~\cite{pinzi2019adaptive, favaro2018automatic, segato2019automated}.
It is difficult to effectively compare these motion planning algorithms since they span different organs and different instances of these organs.

\section{\dbname~}

\dbname~contains anatomical environments and specifications of clinically relevant scenarios in the lungs, liver, and brain. 
These three organs have received considerable research attention from the continuum medical robots motion planning community, especially for steerable needles.
The description of each environment, the clinical motivation, and several relevant evaluation criteria are presented below.
Although each organ has various pathologies each defined by a different clinical objective, the planning problem we consider is target reach.
This objective encompasses many clinical procedures, including biopsy, ablation, and drug delivery.
Future iterations of the data could be adapted to evaluate motion planners for scenarios where the objective is different, such as manipulation at the target.

We represent the environments as three-dimensional binary maps that indicate the presence or absence of obstacles at each corresponding voxel location in the original medical image. 
This environmental representation is used by many of the medical robot motion planners referenced above.
We also provide a collection of clinically-motivated start poses in each environment, along with target points that correspond to true clinical targets.

\subsection{Lungs}

It is estimated that roughly one million pulmonary nodules are discovered every year in the United States.
In order to get a definitive diagnosis for these nodules, a tissue biopsy is required.
There are several methods to reach lung nodules, but the least invasive and safest approach is via bronchoscopy where a physician navigates a bronchoscope through the airways and inserts a needle into the lung tissue towards the target.
Since physicians currently use straight rigid tools to perform the biopsy which are limited in reach and access, there have been efforts to use flexible steerable needles to overcome some of the existing challenges and increase the number of patients for which bronchoscopy can be used~\cite{frieddesign}.
Delivery of the robot can be done through the working channel of a bronchoscope.

\begin{figure}[t!]
    \includegraphics[width=1.0\columnwidth]{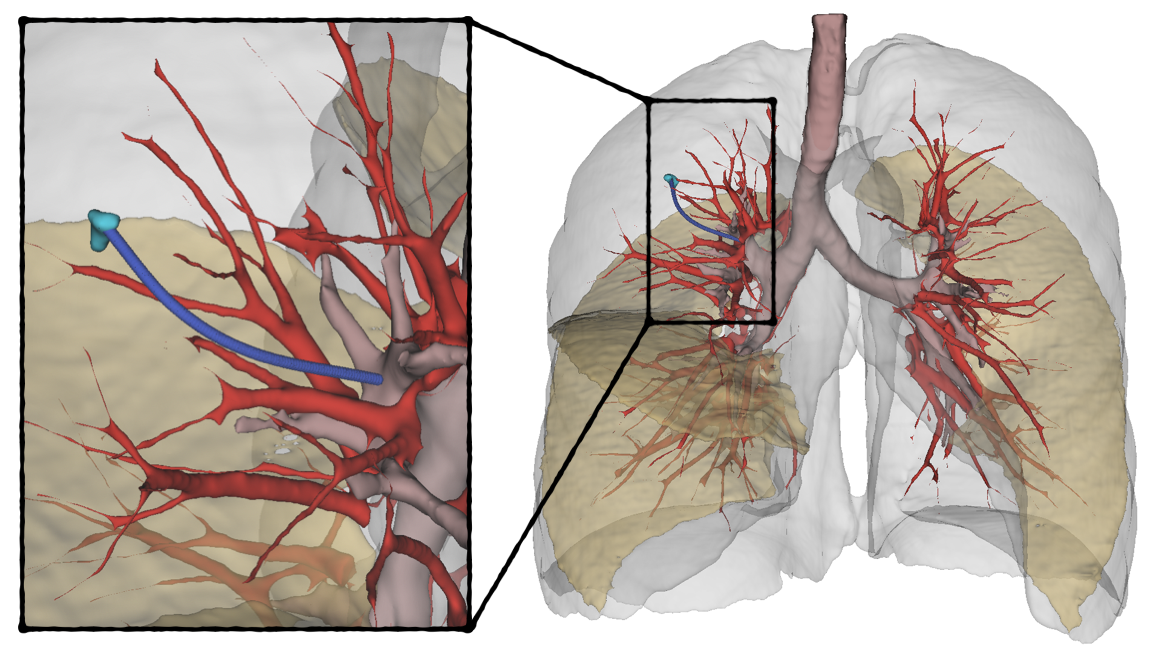}
    \centering
    \caption{
    A representative lung environment consisting of \textcolor{red}{blood vessels}, \textcolor{Goldenrod}{lung fissures}, \textcolor{Salmon}{bronchial tree}, and \textcolor{lightgray}{pleural boundary}. A \textcolor{cyan}{nodule (target)} is shown on the right along with a sample \textcolor{blue}{planned trajectory}. The steerable needle starts at a valid point along the airway wall and can travel through the space within the pleural boundary that is not occupied by obstacles.
    }
    \label{fig:fig1}
\end{figure}

At a high level, the anatomy of the lungs consists of the airways, major blood vessels, and the pleura (lung boundary) (see Figure~\ref{fig:fig1}).
The remaining space inside the lung (known as the parenchyma) is composed of functional tissue and is the location where lung nodules that may be suspicious for cancer often present.
We present 5 motion planning scenarios in the lungs that reflect real clinical scenarios of patients with lung nodules.
The data is part of the 
Lung Image Database Consortium and Image Database Resource Initiative
(LIDC-IDRI) image collection~\cite{armato2011lung, armato2011lungdata} from The Cancer Imaging Archive (TCIA)~\cite{clark2013cancer}.
Each environment consists of obstacles including the airways, major blood vessels, lung fissures, and pleura, and segmented nodules in the parenchyma.
The fissures and nodules were manually segmented while all other objects were automatically segmented~\cite{fu2018safe}.
The start poses correspond to areas along the airway wall that are accessible with a bronchoscope through which a medical robot can be passed.
An example plan for a flexible steerable needle is shown in Figure~\ref{fig:fig1}.

\subsection{Liver}

Currently, percutaneous liver biopsies, where a physician inserts a needle through the abdominal wall and into the liver, are most commonly performed with straight rigid tools.
The mechanical constraint of existing devices make it hard to reach posterior sites that are obstructed by critical anatomy. 
Additionally, when multiple targets exist, a physician will need to re-insert the needle for each target.
Medical devices such as steerable needles that are able to curve around obstacles and reach multiple sites from a single point-of-entry can alleviate some of these clinical challenges.

\begin{figure}[t!]
    \includegraphics[width=0.8\columnwidth]{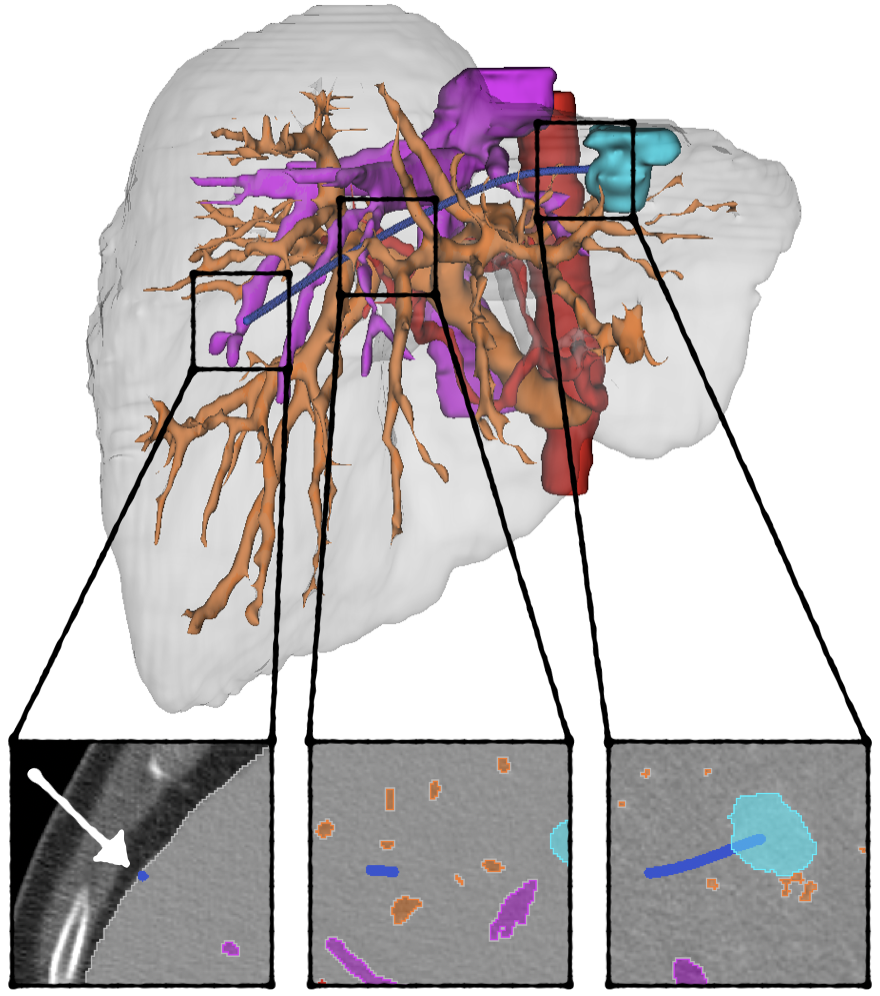}
    \centering
    \caption{
    A liver environment from the dataset showing the segmented \textcolor{red}{hepatic arteries}, \textcolor{Mulberry}{hepatic veins}, \textcolor{orange}{portal vein}, \textcolor{lightgray}{liver boundary}, \textcolor{cyan}{nodule (target)}, and a sample \textcolor{blue}{trajectory}. The three boxes on the bottom show the view in the CT slices (transverse planes).
    }
    \vspace{-4mm}
    \label{fig:fig2}
\end{figure}

Similar to the lungs, the liver motion planning environment consists of major blood vessels and the organ boundary. 
The remaining space within the liver (also referred to as the parenchyma) is traversable.
We present 5 motion planning environments in the liver in patients with hepatocellular carcinoma with segmentations of relevant obstacles.
The data is derived from the Hepatocellular Carcinoma Transarterial Chemoembolization Segmentation (HCC-TACE-Seg) dataset~\cite{morshid2019machine, moawadliverdata} from TCIA~\cite{clark2013cancer}.
A sample liver planning environment is shown in Figure~\ref{fig:fig2}.

\subsection{Brain}

\begin{figure}[b!]
    \includegraphics[width=1.0\columnwidth]{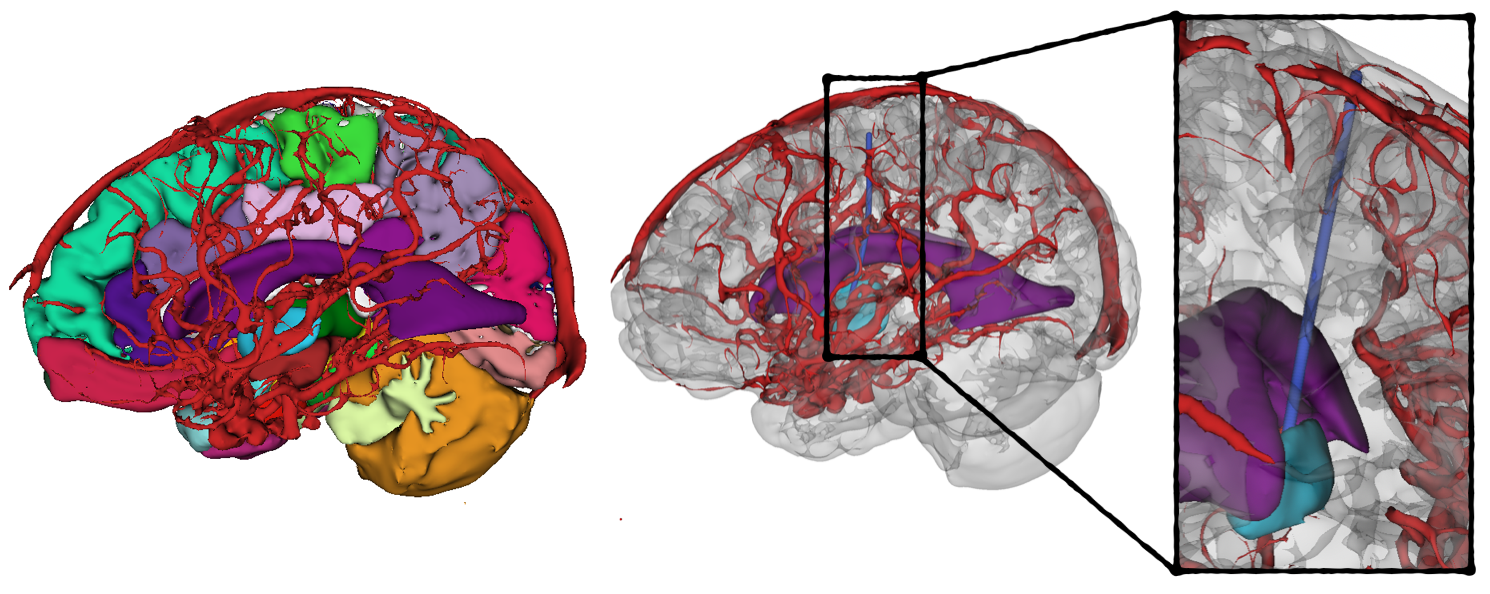}
    \centering
    \caption{
    A brain environment from the dataset showing the segmented \textcolor{red}{blood vessels}, \textcolor{Mulberry}{lateral ventricles} (without the temporal horn), \textcolor{Cerulean}{globi pallidi}, \textcolor{lightgray}{brain boundary}, a sample \textcolor{blue}{trajectory}, and various segmented regions of the brain (left: multiple colors).
    }
    \label{fig:fig3}
\end{figure}

The brain is one of the most complex organs in the body, with nearly every portion of tissue critical to some physiologic function.
From a planning perspective, while obvious obstacles exist such as blood vessels and ventricles, there are many other regions of the brain and properties of the tissue that are important to consider.
For example, the directionality of white matter fibers, which can be analyzed via tractography, can play an important role in evaluating trajectories through the brain. 
Given the density of critical regions in the brain and their fragility, medical robots and motion planning algorithms that consider and account for these constraints can have a large impact in this domain.
We include 5 motion planning environments in the brain where the targets are the globi pallidi for deep brain stimulation.
We consider blood vessels and ventricles as traditional obstacles, whereas all other segmentations of brain regions can be assigned a cost since some subset of them must be traversed.
White matter fiber tracts are not currently included in the data.
The data is part of the Healthy MR Database~\cite{bullitt2005vessel}.
Blood vessels were manually segmented and all other structures were segmented using FastSurfer~\cite{henschel2020fastsurfer}.
A sample environment is depicted in Figure~\ref{fig:fig3}.

\subsection{Evaluation Criteria}

In order to evaluate and compare the performance of different motion planning algorithms, we describe several criteria that are relevant to many medical applications.
This is a general and non-exhaustive list of relevant criteria, and in many cases, each organ and clinical application has domain-specific considerations that would be valuable to use as evaluation metrics.
The following metrics can be reported for a single clinical target or as a statistic across a collection of clinical targets in the data.

\begin{itemize}
    \item \textit{Path Length}: the length of the collision-free motion plan from the start pose to the target.
    \item \textit{Computation Time}: the amount of computation time that the motion planner took to find a kinematically-feasible collision-free plan from the start pose to the target prior to the procedure.
    \item \textit{Replanning Time}: the amount of time that the motion planner took to find a kinematically-feasible collision-free plan from its current intraoperative pose to the target following a random deviation event.
    \item \textit{Obstacle Clearance Statistics}: the minimum, mean, and median of the Euclidean distances between every pose along the motion plan to its nearest obstacle.
    \item \textit{Procedure Success (clinical targets)}: the percentage of clinical targets that the motion planner was able to successfully plan to.
    \item \textit{Coverage of the Anatomy (random targets)}: the percentage of random goals that the motion planner was able to successfully plan to. This is an approximation for the motion planner’s ability to generalize to any target.
\end{itemize}
\section{Discussion}

In this work, we proposed \dbname, a new medical benchmarking dataset for motion planners for medical continuum robots and related minimally invasive medical robots.
At its current state, \dbname~is a stand-alone collection of clinically-relevant motion planning scenarios.
It is our hope to extend the benchmarking suite in the various ways described throughout this paper, as well as to integrate it directly into an existing motion planning framework.
We also hope to expand the provided start poses to include start regions from which motion planning can begin.
This would introduce an interesting and medically relevant problem where the choice of start pose is itself a motion planning challenge that can be optimized as part of the procedure.
We also hope to introduce uncertainty into the data by implementing methods that allow for target and obstacle deviation during a medical procedure.
Uncertainty is highly likely to have an impact during a procedure because of the deformable nature of organ tissue.
Ideally, the environmental uncertainty would be incorporated into a simulator that would consider a robot's model and differential constraints to enable more realistic evaluations.
Since the data is not directly tied to medical continuum robots, the clinical environments can be used to evaluate other existing and novel medical robots.
It is our intention that this dataset be used to advance research in motion planning for autonomous medical robots towards the ultimate goal of leveraging these systems to improve patient care.

\bibliographystyle{IEEEtran}
\bibliography{references, bibabbrevs}

\end{document}